\documentclass[10pt]{article} 

\usepackage{balance} 
\usepackage[english]{babel}
\usepackage[utf8]{inputenc}
\usepackage{subcaption}
\usepackage{multirow}
\usepackage{multicol}
\usepackage{amsmath}
\usepackage{mathtools}

\usepackage{enumitem}
\usepackage{siunitx}
\usepackage{graphicx}
\usepackage{fullpage}
\usepackage{natbib}
\usepackage{booktabs}
\usepackage{url}
\usepackage{subcaption}

\title{School of hard knocks: Curriculum analysis for Pommerman with a fixed computational budget}

\author{Omkar Shelke$^1$, Hardik Meisheri$^1$, Harshad Khadilkar$^{1,2}$\\$^1$TCS Research, Mumbai, India\\$^2$IIT Bombay, Mumbai, India}

\newcommand{\BibTeX}{\rm B\kern-.05em{\sc i\kern-.025em b}\kern-.08em\TeX}

\begin{document}



\date{}
\maketitle 

\begin{abstract}
	
	Pommerman is a hybrid cooperative/adversarial multi-agent environment, with challenging characteristics in terms of partial observability, limited or no communication, sparse and delayed rewards, and restrictive computational time limits. This makes it a challenging environment for reinforcement learning (RL) approaches. In this paper, we focus on developing a curriculum for learning a robust and promising policy in a constrained computational budget of 100,000 games, starting from a fixed base policy (which is itself trained to imitate a noisy expert policy). All RL algorithms starting from the base policy use vanilla proximal-policy optimization (PPO) with the same reward function, and the only difference between their training is the mix and sequence of opponent policies. One expects that beginning training with simpler opponents and then gradually increasing the opponent difficulty will facilitate faster learning, leading to more robust policies compared against a baseline where all available opponent policies are introduced from the start. We test this hypothesis and show that within constrained computational budgets, it is in fact better to ``learn in the school of hard knocks'', i.e., against all available opponent policies nearly from the start. We also include ablation studies where we study the effect of modifying the base environment properties of ammo and bomb blast strength on the agent performance.
	
\end{abstract}


\section{Introduction} \label{sec:intro}

Reinforcement learning (RL) has shown remarkable results in gaming environments with complex decision-making, such as Atari, Dota2 and Starcraft. The decisions in these environments range from simple cardinal moves to developing and executing long-term strategies. However, these approaches are typically designed with extremely large computational resources in mind. Applying deep reinforcement learning (DRL) to new and niche games with a fresh and complex mix of challenges is an opportunity to introduce sample efficiency goals into RL. These challenges include partial observability, imperfect information, highly dynamic environment, long and sparse rewards, and credit assignment. Most of these challenges are still open problems by themselves, and when these are combined into an environment such as Pommerman, require many innovations in applications of RL. 

Pommerman is a 2-\textit{vs}-2 game environment introduced in 2018 (see Section \ref{sec:environment} for details), and is a modified version of the popular game `Bomberman'. It (and gaming environments in general) is a good testbed for RL, because the environment interactions are compact and relatively inexpensive, but also include the challenges described above. A common method for improving the sample efficiency of RL is to seed the policy using imitation learning (IL), which involves imitating (in a supervised manner) an expert policy. We do not have the luxury of having known expert or optimal policies in the case of Pommerman; however, a prior baseline provided by default with the environment is still useful for learning the basics of the game. We seed all our RL agents with the same imitation-learnt policy, as described in Section \ref{sec:method}.

A majority of prior literature supports the use of a \textit{curriculum} for training reinforcement learning algorithms to solve complex environments, with the emphasis being on learning basic skills on simpler versions of the environment, before graduating to handle the full complexity. However, this notion could result in a longer training run, since one should ideally wait for learning to stabilise at each level of complexity, before introducing the next level. In this paper, we consider the problem of \textit{designing an effective curriculum within a fixed computational budget}. We do this because (i) not many institutions have access to unlimited time and hardware capacity for training, and (ii) even if such capacity is available, the benefit of resulting policy improvements may be outweighed by the cost of producing the improvements. Thus we ask the question:

\textit{``Given a fixed computational budget of 100,000 games and starting from a policy trained to mimic a simple baseline, what is the best curriculum for training a given reinforcement learning algorithm to play against a specified set of opponent policies?''}

A curriculum is an efficient tool for humans to progressively learn from simple concepts to hard problems. It breaks down complex knowledge by providing a sequence of learning steps of increasing difficulty. To aid in learning skills, we also define a novel reward function which helps in mitigating issues arising from sparse rewards. Pommerman is a complex environment where the required skill-sets might depend on the different opponents and strategies they employ. Therefore, it was our hypothesis that introducing different mixes of opponent strategies at various points in the training run would have an effect on the robustness of policies learnt. We defined the computational budget in terms of number of games, because the number of frames (or time steps) can vary significantly depending on self and opponent policies. Furthermore, different game situations result in different response times from the environment and the agent(s). As a result, defining a limit of 100,000 games seemed as good as any other metric. We also note that this budget is significantly short (10\% or less) of the number of games reported by most studies on Pommerman in literature.

We believe that the major contributions of this paper, are, (i) investigating curriculum learning in limited budget settings for multi-agent settings, (ii) proposing new reward structure for complex games such as Pommerman based on approximate skill-sets, and (iii) extensive experimentation and ablation studies to provide improved understanding about learned policies.


\section{Related Work} \label{sec:related}

To solve a complex problem such as Pommerman, a popular method is to use expert demonstrations and learn with behavioural cloning or imitation learning~\cite{meisheri2019accelerating}. These methods explicitly encourage the learned policy to mimic an expert policy \cite{bain1995framework, ross2011reduction, daume2009search, zhang2016query, laskey2016shiv, nair2018overcoming, hester2017deep, ho2016generative, aytar2018playing, lerer2018learning, peng2018deepmimic}. However, these imitation policies are shown to diverge and perform poorly when they encounter new states, and they suffer from compounding error problem when using a neural network as a policy function approximator~\cite{ross2011reduction}. Furthermore, many such studies solve the problem in a single agent environment without partial observability and with near-optimal expert policies. In our case, we do not have access to such trajectories and moreover, the environment is multi-agent, partially observable, and highly dynamic. Hence we use imitation learning from noisy expert data generated by the default heuristic agent provided by the environment. Apart from serving as a proxy for the expert, it also provides diversity in the distribution of state space \cite{meisheri2019accelerating}.  

Tree-based techniques such as Monte Carlo Tree Search (MCTS) have been shown to perform well in Pommerman~\cite{pomm_book_chap, osogami2019real, zhou2018hybrid}. However, they require much more computational infrastructure than pure RL, in addition to significant human effort for evaluating the trajectories. In the learning agent category of the 2018 Pommerman competition at NeurIPS, Navocado was the best agent~\cite{peng2018continual}. It used an A2C agent and continual match based training (COMBAT) training framework. They showed that during the course of training, the agent learnt different skills such as hiding and placing bombs. While this shows the value of using curriculum learning, they used a distributed learning framework which required large computational requirements (220 CPU cores and 32 GPUs). In addition, agents were trained against only one policy (heuristic provided from competition) which can lead to overfitting. Other studies \cite{gao2019skynet} show an instance where an agent learns to stand diagonally opposite to the opponent, and the default heuristic agent is unable to move (even though moves are available). This was shown to be a bug in the heuristic, and is obviously not replicable in other policies. One prior study \cite{meisheri2019accelerating} also uses a curriculum after imitation learning, significantly reducing the training time. However, they also train against only one agent policy. We delve deeper into understanding what would be a better training paradigm in limited budget setting for a generalizable agent policy against a mix of opponents.


\section{Pommerman Environment} \label{sec:environment}

\begin{figure}
    \centering
    \includegraphics[width=0.4\textwidth]{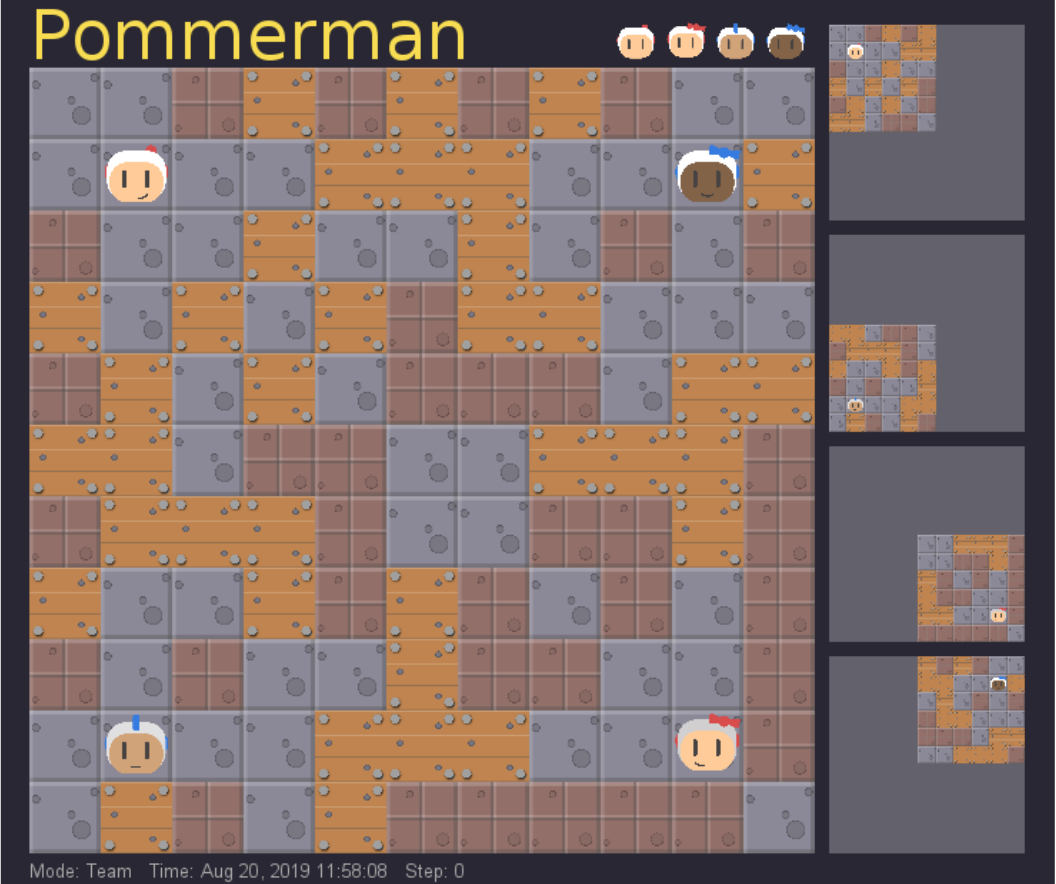}
    \caption{Pommerman initial state. Observations visible to each agent are shown on the right.}
    \label{fig:board}
\end{figure}

Our work focuses on the Pommerman Team Environment which comprises of a partially observable 11 $\times$ 11 grid with four agents (two agents in each team) where teammates start from opposite corners\footnote{This environment was used for the NeurIPS 2018 competition.} as shown in Figure~\ref{fig:board}. Each agent has a limited visibility of 5 grid cells in each direction. The board comprises of wooden walls, rigid walls, and passages. Rigid walls are indestructible and impassable, whereas wooden walls can be destroyed by bombs. The objective of the team is to win the game by killing its opponents by placing bombs. If the game does not end before the maximum timesteps (800) or if all living opponents die at the same timestep, the result is a tie. Bombs explode only in the 4 cardinal directions (shape of a vertical cross) with a default blast radius of $2$. An agent can pick up different power-ups during the game, obtained by blasting wooden walls (50\% of the wooden walls have powerups behind them). There are three kinds of power-ups, which (i) increase the agent's ability to place multiple bombs (\textit{Extra Bomb}), (ii) increase blast radius (\textit{Increase Range}), and (iii) ability to kick away a nearby bomb (\textit{Can Kick}). In each timestep, the agents can choose from 6 discrete actions, which are \textit{Stop}, \textit{Up}, \textit{Left}, \textit{Down}, \textit{Right}, \textit{Bomb}. The \textit{Bomb} action lays a bomb, the \textit{Stop} action has no effect on the agent's position, and the remaining four actions are for movement. Each game starts with a random generation of rigid and wooden walls which are symmetric along the principal diagonal, in a formation that ensures all agents have accessible paths to each other. 

\section{Proposed Methodology} \label{sec:method}

\subsection{Preliminaries}
In the Pommerman team setting, the agents participate and interact with the environment concurrently. Each agent can observe a specific part of the state, while the rest of the state and other agents' actions are not known. This kind of environment can be formulated as Partially Observable Markov Decision Process (POMDP)~\cite{littman1994markov},  $(\mathcal{S}, \mathcal{A}, \mathcal{T}, \mathcal{R}, \Omega, \mathcal{O})$ where $\mathcal{S}$ represents the set of states, $\mathcal{A}$ denotes the set of actions, $\mathcal{T}$ represents the state transition function, $\mathcal{R}$ denotes the reward function, $\Omega$ denotes the set of observations an agent would experience, and $\mathcal{O}$ is observation function $\mathcal{O}: \mathcal{S} \times \mathcal{A} \rightarrow \mathcal{P}(\Omega)$ which provides probability distribution for observations in the next step, given current state and action~\cite{kaelbling1998planning}. This formulation can be scaled to multi-agent systems such as Decentralized POMDP (Dec-POMDP). Given a team of $n$ agents, $\mathcal{S}$ is a finite set of states $(s_1,\ldots, s_m), \mathcal{A}$ is set of joint actions for agents 1 to $n$, with individual action spaces defined by $\mathcal{A}_1,\ldots, \mathcal{A}_n$. An instance of the joint action is represented as $(a_1,\ldots, a_n)$. $\mathcal{T}(s,(a_1,..., a_n), s')$ is the transition function, $(\Omega_1,..., \Omega_n)$ are the set of observation for agents 1 to $n$. The agents receive a single, joint reward $\mathcal{R}(s, a_1,..., a_n, s')$ for the team. However, such a formulation leads to a non deterministic exponential time complete problem when n > 2, as shown in~\cite{bernstein2002complexity}. Therefore we model the present problem as a POMDP assuming all other agent policies are stationary. 
Proximal Policy Optimization (PPO)~\cite{schulman2017proximal} has been successfully applied to Pommerman~\cite{meisheri2019accelerating, gao2019skynet, meisheri2020sample} with significant improvement over other learning agents such as A2C~\cite{peng2018continual}. However, it is prone to sample inefficiency in its vanilla form. We utilize imitation learning and reward shaping to reduce the sample inefficiency and speed up the training.

\subsection{Reward Shaping}

In a multi-agent environment with partial observability and sparse rewards, assigning credit among teammates is a major challenge in the application of learning-based algorithms. In addition, there are many ways for games to be tied or lost (timeout, all agents died, one opponent dead, and so on). A single terminal reward is also insufficient when determining the contribution of each agent to the final outcome. For instance, consider a game where an agent eliminates an opponent and then dies, after which its teammate eliminates the remaining opponent. Under this scenario, both team members get a positive reward from the environment, but this could also reinforce the mistake made by the first agent. Similarly in another instance, one agent could eliminate both opponents whereas its teammate hides in a different part of the board; both agents would get positive rewards, reinforcing a lazy agent policy \cite{panait2005cooperative}. Reward shaping has been used in RL to overcome these problems~\cite{ng1999policy, dorigo1994robot, lample2017playing, song2019playing, OpenAI_dota}, but can also lead to corruption of true value functions and can sometimes lead to undesirable policies being reinforced~\cite{FaultyReward, reward_bair}. This is especially true if the dense rewards are approximated from partially observed states. With these competing considerations in mind, we subtly redefine the rewards as given in Table \ref{tab:reward_shaping}.

To overcome the credit assignment problem, we have developed a tracking module which measures the acquisition of agent skills and provides intermediate rewards over the course of the game. These skills include blasting of wooden walls, picking of powerups, exploration of map, and killing of opponents, and the rewards are specifically based only on the observations available to the agent. The cumulative reward signal provided to the agent $i$ is given by,
\begin{equation}
    R_i(t) = \underbrace{0.5\,\times\,E_i}_{Environment} + \underbrace{0.5\,\times\,K_i}_{Kill} + \underbrace{T_i(t)}_{Intermediate}  
    \label{eq:reward}
\end{equation}
where $E_i$ is the default environment reward, $K_i$ is a specific \textit{kill} reward (given if the opponent agent is killed in the playing agent's visible state space and by its bomb only), and $T_i$ is the intermediate reward. $E_i$ and $K_i$ are available at the terminal step and $T_i$ is evaluated per timestep. They are described in Table \ref{tab:reward_shaping}. 

The tracking module keeps track of the bomb placed by the agent until it blasts and the damage caused by the bomb (\textit{number of wooden walls blasted, opponents killed, teammate killed, suicide committed}). The module observes the changes in the agent's visible state and focuses on self-actions. There are certain edge cases which are difficult to judge, such as when an agent places a bomb at a certain location and then moves such that the bomb is no longer in the visible part of the state. In such cases, the tracking module does not provide a kill reward (since credit cannot be proven). We also remove any kill reward if the bomb is kicked in any direction. 

\begin{table}
    \caption{Rewards per agent (all algorithms).}
    \label{tab:reward_shaping}
    \centering
    \begin{tabular}{l S[table-format=3.2]} \toprule
    \textit{Description} & \textit{Intermediate Reward} \\ \midrule
    Visiting a new unique cell & 0.01 \\ 
    Picking any powerup & 0.01 \\
    Blasting $n$ wooden wall cells & 0.01$\,n$ \\ \midrule
     & \textit{Kill Reward} \\ \midrule
    Death of an opponent & 0.50 \\
    Death of a teammate & -0.50 \\
    Own death & -1.00 \\ \bottomrule
    \end{tabular}
\end{table}

\subsection{Training} \label{subsec:training}

\begin{table}[h]
    \caption{Opponent agent summary. Smart Simple and Smart Simple NoBomb agents are based on the default SimpleAgent provided with the environment, but include action filtering to block definitely suicidal actions in a given state.}
    \label{tab:opponent_names}
    \centering
    \resizebox{\linewidth}{!}{%
    \begin{tabular}{lll} \toprule
    \textit{Acronym} & \textit{Name} & \textit{Description} \\ \midrule
        ST & Static & Agent which does not move from its initial position, takes only \textit{stop} action \\ 
        SS & Smart Simple & Simple Agent with action filter \\
        SS-NB & Smart Simple NoBomb & Smart Simple Agent with bomb action forbidden \\
        PPO-18 & PPO\_2018 & Agent learned using imitation from simple agent followed by PPO ~\cite{meisheri2019accelerating}  \\ \bottomrule
    \end{tabular}
}
\end{table}

The training is accomplished in two phases: an imitation learning phase followed by reinforcement learning. During the RL training phase, we make use of a curriculum along with a reshaped reward function as described above. Imitation learning approach was adopted from \cite{meisheri2019accelerating}, where we collect samples from the default agent (SimpleAgent) provided with the environment, over 50K games, and then train the policy network using supervised learning. The imitation learnt model acts as the initial policy network during the RL training phase (with PPO). We avoid using any regularization technique such as dropout while training using PPO, as this leads to significant increase in KL divergence during training (too high rate of change in policy).

A \textbf{curriculum} is meant to optimize the order in which experiences are accumulated, so as to enhance the performance or training speed on a set of tasks. Given a set of samples, sequencing of opponent policies can be utilized in a way that facilitates learning. We define the curriculum by picking an order of opponents to train against. The order of opponents is decided on the basis the opponent complexity and skillset focus. We gradually increment the degree of opponent difficulty over the course of the fixed RL training budget of 100k games for attaining a robust and generalized policy effectual against each opponent (listed in Table \ref{tab:opponent_names}).  

At the start of RL training, the policy network is initialised with the pre-trained IL weights. The value network is initialized using random weights, and then the following curriculum is followed for training with PPO over 100k games. For all agents listed in Table \ref{tab:trained_agent_names}, the first 5k episodes run with only value network training (policy network frozen). This step ensures that both the policy and value networks are brought to reasonably trained levels. Following this, the six agents follow a phased curriculum. Agent$_0$ is the baseline PPO agent that trains against all 4 opponents equally from the start, for the entire remaining budget of 95k games. Agent$_{20}$, Agent$_{40}$, and Agent$_{60}$ train against only the static opponent for 20k, 40k, and 60k games respectively, followed by mixed training against all opponents. The last two agents train in 4 phases each. Agent$_{focus}$ eventually trains against each opponent for the same number of episodes as Agent$_0$, but it trains against only one opponent at a time. Finally, Agent$_{incrm}$ also trains against each opponent for a similar number of games (between 23k and 24k against each opponent), but it follows a curriculum with incremental difficulty.

\begin{table}
    \centering
    \caption{Naming convention for different agents trained, and their curriculum. Opponent acronyms can be correlated with Table \ref{tab:opponent_names}. A total budget of 95k is available, since all agents train their value networks for the first 5k games against all opponent types.}
    \label{tab:trained_agent_names}
    \begin{tabular}{l|c|cccc} \toprule
	    \textit{Name} & \textit{Phase} & \multicolumn{4}{c}{\textit{Opponents}} \\
	    &  & ST & SS & SS-NB & PPO-18 \\
	    \midrule 
        Agent$_{0}$ & 1 & 23.75k & 23.75k & 23.75k & 23.75k \\ 
        \hline
        \multirow{ 2}{*}{Agent$_{20}$} & 1 & 20k & 0 & 0 & 0 \\ 
         & 2 & 18.75k & 18.75k & 18.75k & 18.75k \\ 
        \hline
        \multirow{ 2}{*}{Agent$_{40}$} & 1 & 40k & 0 & 0 & 0 \\ 
         & 2 & 13.75k & 13.75k & 13.75k & 13.75k \\ 
        \hline
        \multirow{ 2}{*}{Agent$_{60}$} & 1 & 60k & 0 & 0 & 0 \\
         & 2 & 8.75k & 8.75k & 8.75k & 8.75k \\
        \hline
        \multirow{ 4}{*}{Agent$_{focus}$} & 1 & 23.75k & 0 & 0 & 0 \\
         & 2 & 0 & 23.75k & 0 & 0 \\
         & 3 & 0 & 0 & 23.75k & 0 \\
         & 4 & 0 & 0 & 0 & 23.75k \\
        \hline
        \multirow{ 4}{*}{Agent$_{incrm}$} & 1 & 6k & 0 & 0 & 0 \\
         & 2 & 5.8k & 8k & 0 & 0 \\
         & 3 & 6k & 8k & 11.6k & 0 \\
         & 4 & 6.2k & 7.8k & 11.6k & 24k \\
        \bottomrule
    \end{tabular}
\end{table}
For all agents except Agent$_0$, the reason for initially playing exclusively against a static (ST) opponent is to learn basic rules and skills in the environment, without complication from opponent policies. Since ST is a stationary agent, its opponent needs to venture out and hunt for it in order to register a successful win. As opponents of higher complexity are added, the hypothesis is that higher-level skills will be learnt in the final phases. The Smart Simple (SS) agent is the default Pommerman agent but with an action filter for avoiding decisions that cause certain death. However, the underlying heuristic is highly aggressive and occasionally traps itself with bomb placement. This is why the SS-NB agent (which is the same as SS but without bomb-laying capability) is harder to defeat; it has the intelligence to defend itself without the propensity to commit suicide. The last two agents in Table \ref{tab:trained_agent_names} are designed to have the same total experience as Agent$_0$, but with a different curriculum. This should allow us to tease out the specific effect of the curriculum on the training, while Agent$_{20}$ to Agent$_{60}$ will allow us to estimate the value of learning basic skills.

It was reported in \cite{meisheri2019accelerating} that training directly against a challenging opponent leads to the development of a cautious agent (passive, fleeing from danger). In addition, such training affects the basic game-playing skills like the placing of bombs which is crucial for winning the game. On the other, training purely against ST is not helpful, since our agent will not learn complex tactics such as trapping, bomb avoidance, or the use of powerups. Therefore, we try to find a balance between the training budget expended on ST versus training against all the opponents in order to absorb the skills in a robust manner. 

\subsection{Experimental setup} \label{sec:experiments}

We generate trajectories by playing 64 parallel games between training cycles, giving us 128 agent trajectories (a team consists of 2 agents using the same policy weights). We train in a single batch aggregated over 128 trajectories with a learning rate of $0.001$. Discount factor $\gamma$ is set to $0.99$ and the generalized advantage estimate factor $\lambda$ is set to $0.95$ during all the PPO training runs.

As described in Section \ref{subsec:training}, we start all agents from the same imitation-learnt policy, followed by 5k games of value network training. This is followed by different curricula as given in Table \ref{tab:trained_agent_names}. We simulate and collect trajectories by playing against the opponents as per the experiment specification. 
The model learns from the reshaped reward (\ref{eq:reward}) evaluated during trajectory roll-outs.



\section{Basic Results} \label{sec:results}

In this section, we describe the key observations during the training and evaluation phase. Ablation studies based on the observations are described in the next section.

\subsection{Training}

\begin{figure}[h]
	\centering
	\includegraphics[width=0.9\textwidth]{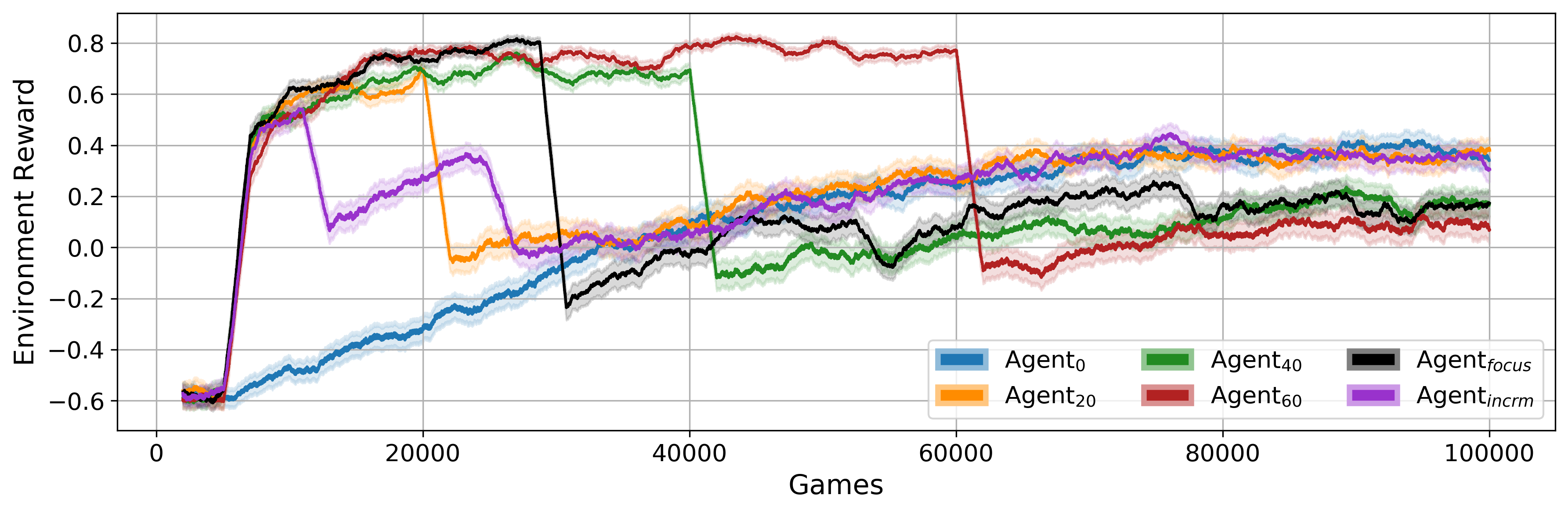}
	\caption{Training over $100$k games, where y axis denotes default environment reward.}
	\label{fig:env_rwd}
\end{figure}

Figure \ref{fig:env_rwd} summarises the entire training process for all six agents (see Table \ref{tab:trained_agent_names}) over the 100k computational budget. All the plots are generated using a moving average window of $2000$ games and the shaded regions denote $95\%$ variance bounds within the moving window. Note that the y-axis denotes the default environment reward and not the reshaped version. All agents show similar performance for the initial 5k games, when only the value network is being trained. After this, all the agents except Agent$_0$ show a sharp increase as they begin to play against the ST (static) opponent. Agent$_0$ is the exception since it plays against all opponents from the start. The staggered introduction of various opponents is marked by sharp changes in reward values for all the curves. We see that Agent$_0$, Agent$_{20}$, and Agent$_{incrm}$ converge in the same range and are better than the rest.

\begin{figure}
	\centering
	\includegraphics[width=0.5\textwidth]{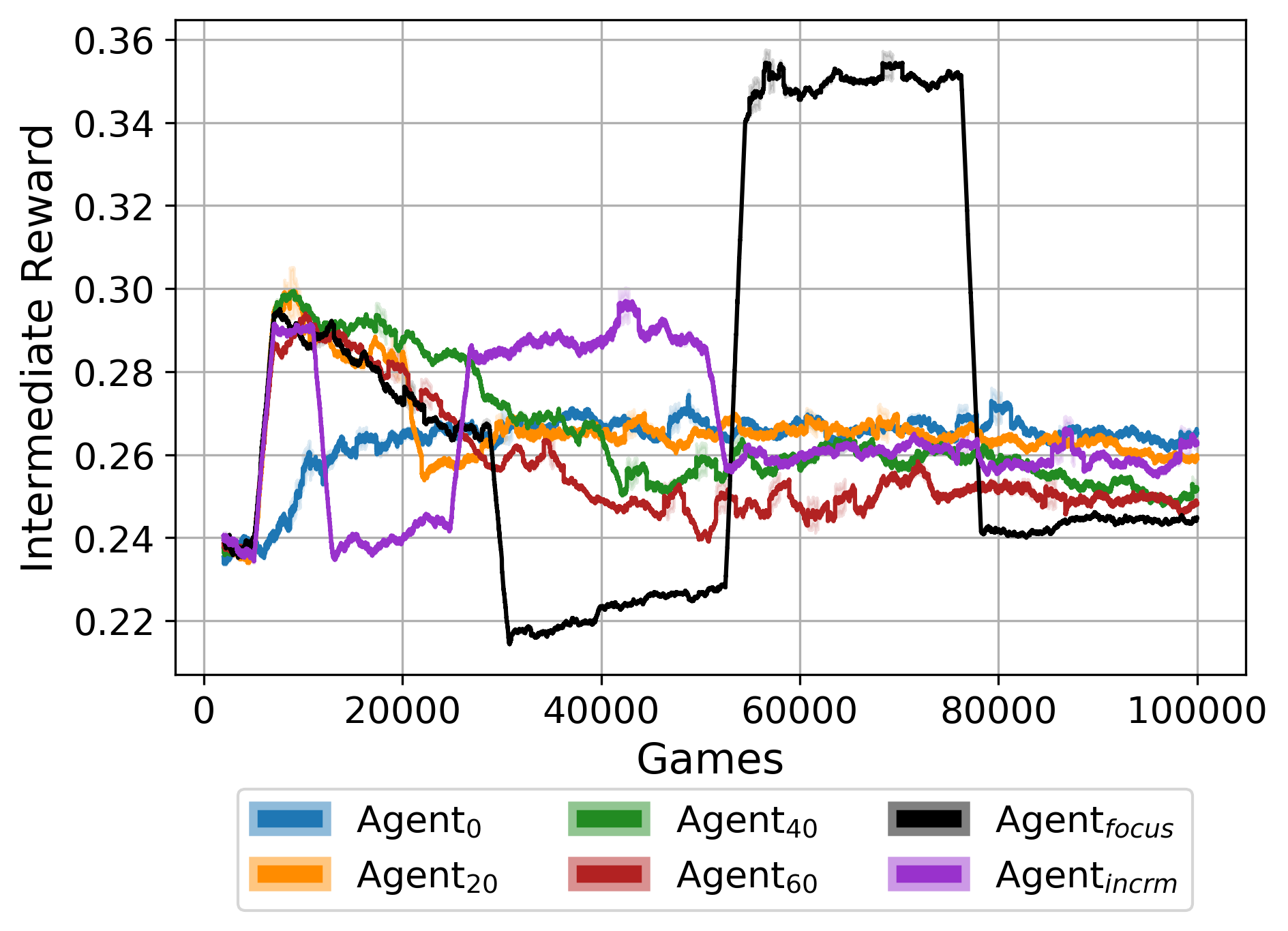}
	\caption{Intermediate reward during training (plotted with a moving average of $2$k games)}
	\label{fig:intermediate_rewards}
\end{figure}

The intermediate reward $T_i(t)$ generated from the  tracking module (see Table \ref{tab:reward_shaping}) is plotted in Figure ~\ref{fig:intermediate_rewards}. We note that the sharp spike between 50k and 80k games for Agent$_{focus}$ is during training against SS-NB exclusively. Since this opponent is purely defensive, the game lengths are typically longer, giving the agent an opportunity for more exploration and powerups. However, we also observe that this intense focus on intermediate rewards eventually results in poor rewards at the end. By contrast, Agent$_{40}$ and Agent$_{60}$ are at the other extreme, where they spend a long time training against ST. This results in a highly aggressive policy that directs the agent to the opponent location, ignoring all other considerations. There is a resultant degradation in the intermediate rewards from which the agents do not seem to recover. Agent$_0$, Agent$_{20}$, and Agent$_{incrm}$ seem to strike the right balance between intermediate rewards and actually killing the opponent. 

\begin{figure}[h]
	\begin{subfigure}{.5\textwidth}
		\centering
		\includegraphics[width=0.95\textwidth]{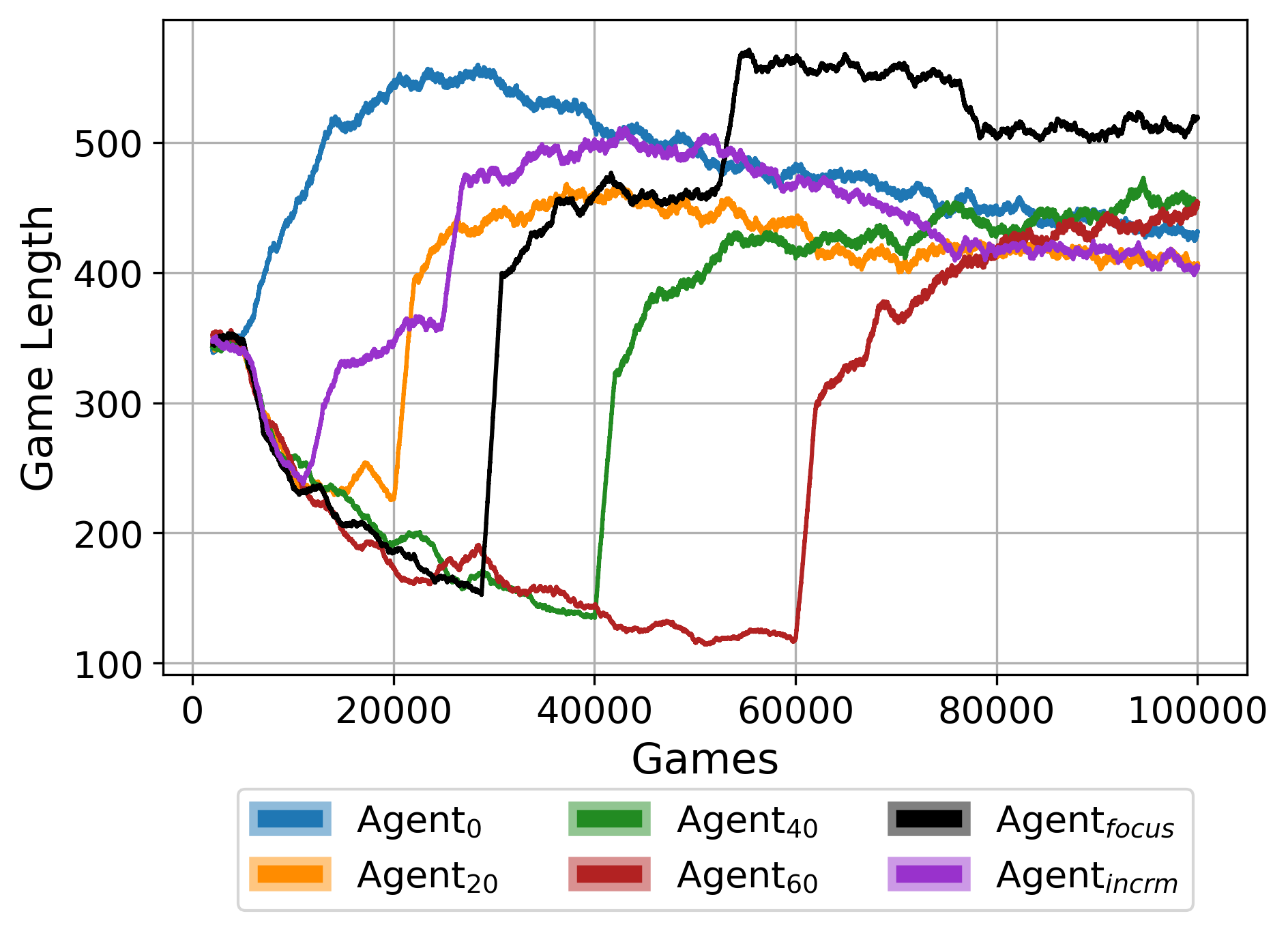}
		\caption{Game length during training (plotted with a moving average of $5$k games)}
		\label{fig:game_length}
	\end{subfigure}%
	\begin{subfigure}{.5\textwidth}
		\centering
		\includegraphics[width=0.95\textwidth]{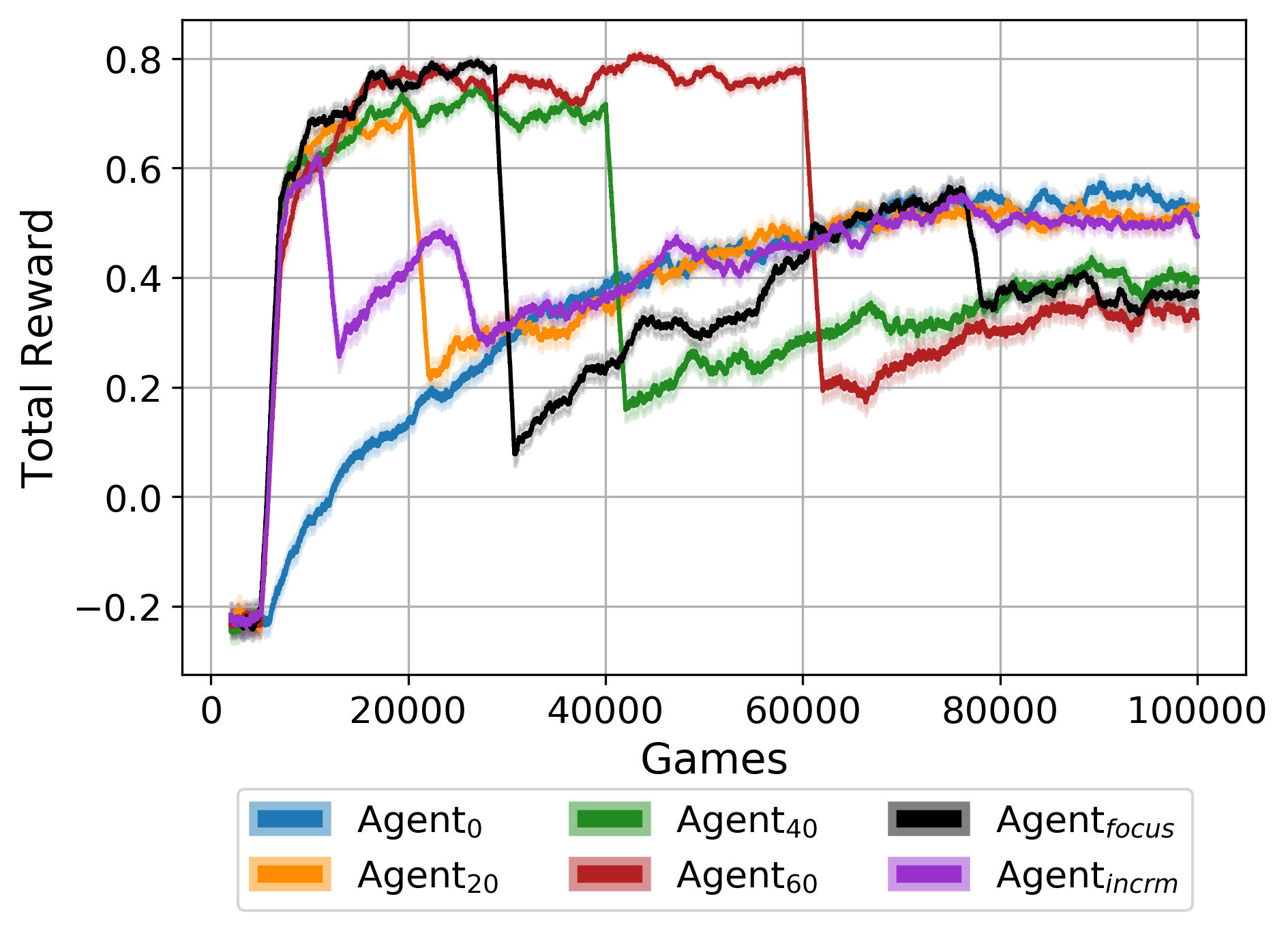}
		\caption{Total reward during training (plotted with a moving average of $2$k games)}
		\label{fig:total_rwd}
	\end{subfigure}%
\end{figure}

Figure~\ref{fig:total_rwd} is in agreement with this conclusion, where these three agents achieve the highest total rewards $R_i(t)$ as defined in (\ref{eq:reward}). Figure ~\ref{fig:game_length} shows the game length of the different experiments. A sharp decrease in game length is observed for versions that are training against only ST, because the opponent location is known and the opponent is static. However, this determinism also diminishes the skills that the agent acquired during imitation learning (exploring the environment, picking powerups, avoiding bombs placed by other agents). The resultant policy degradation results in poor performance when other opponents are introduced. 

\subsection{Post-training evaluation}

Table \ref{tab:ammo1_results} shows the win, loss and tie percentage against all the opponents for a total of 2k games (500 against each opponent). The salient conclusions from this table are,
\begin{itemize}
    \item Agent$_{focus}$ has the best win ratio against ST, and with Agent$_0$ also has no losses against this opponent. This is a little counter-intuitive, because these agents have the lowest number of games against ST during training.
    \item Agent$_{20}$ has the best win ratio against SS and SS-NB. In fact, this agent has a top-3 win ratio across all opponents, and we recall that it is also among the three best performing agents during training.
    \item Agent$_{incrm}$ performs the best against the toughest opponent in this study, PPO-18.
\end{itemize}
%


\begin{figure}
	\begin{subfigure}{.5\textwidth}
		\centering
		\includegraphics[width=0.95\textwidth]{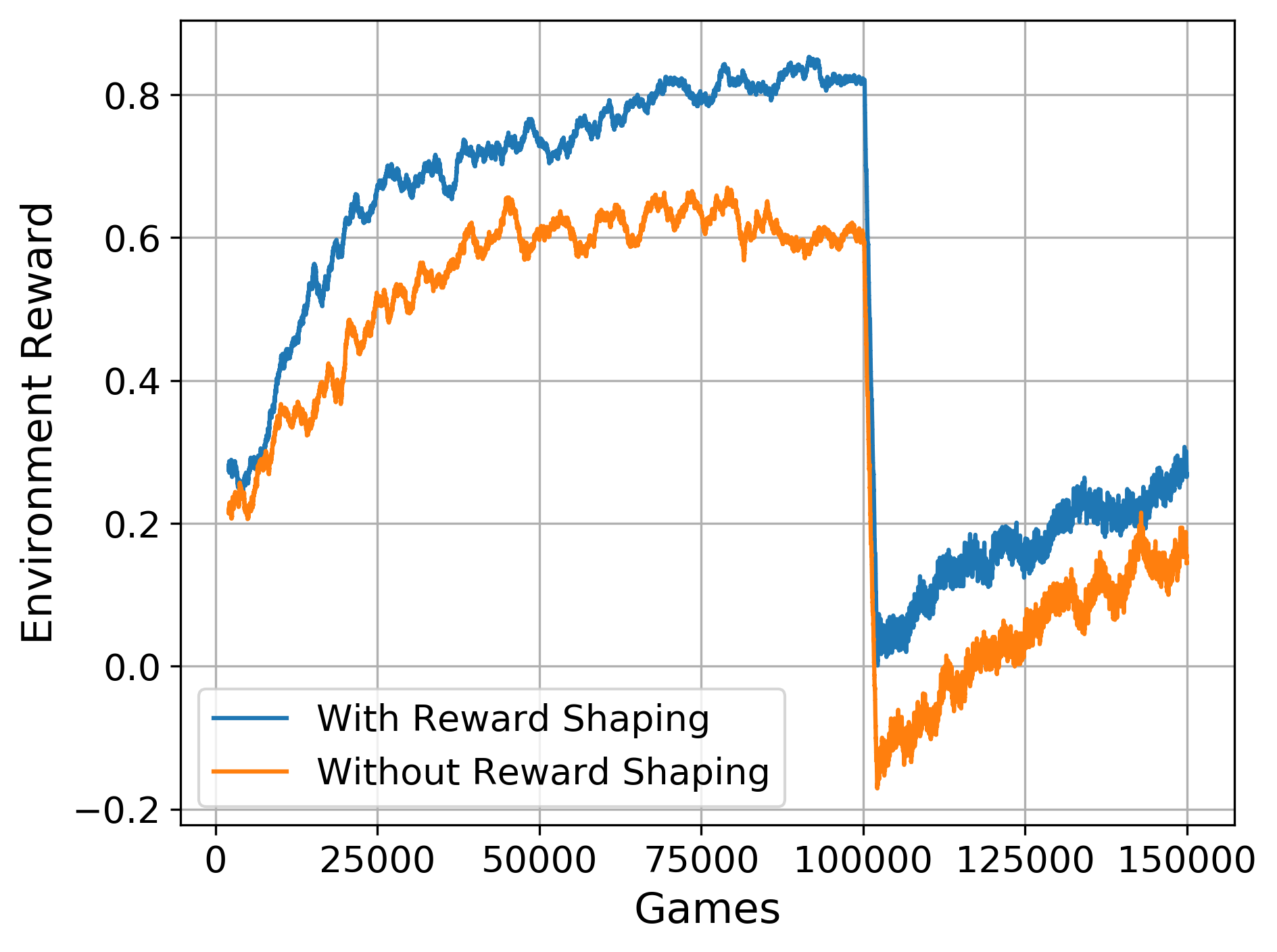}
		\caption{Environment reward progression with and without the structured reward for $150$k games ($100$k - StaticAgent, $50$k - Against all opponents)}
		\label{fig:reward_shaping}
	\end{subfigure}%
	\begin{subfigure}{.5\textwidth}
		\centering
		\includegraphics[width=0.95\textwidth]{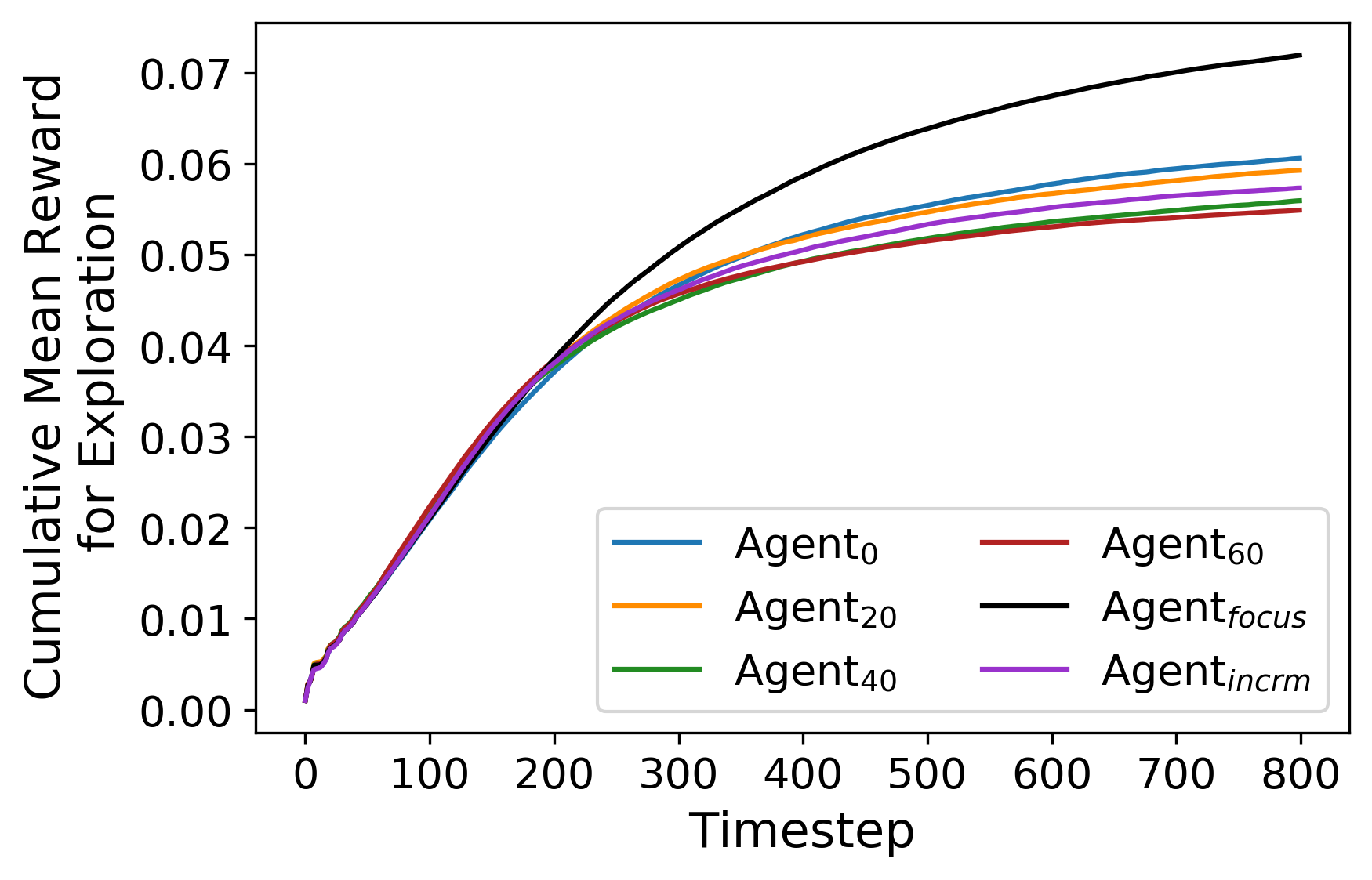}
		\caption{Exploration reward in default environment settings for all types of model variants}
		\label{fig:agent_noch_exploration_rwd}
	\end{subfigure}%
\end{figure}

In order to understand the counter-intuitive result against ST, we plot the exploration reward for all agents in Figure \ref{fig:agent_noch_exploration_rwd}. As described in Table \ref{tab:reward_shaping}, this reward is received for each new cell visited by an agent. Figure \ref{fig:agent_noch_exploration_rwd} averages the reward over all 2k evaluation games and both teammates. We note that Agent$_{focus}$ has the highest exploration among all agents. This might allow it to find paths to its static opponents even when they are hidden behind a complex wall structure, thus converting a larger number of ties into wins.






\begin{table}
    \caption{Evaluation results averaged across $2000$ games for each opponent, values represents $\%$ of wins, loss and ties. Average values are computed only over learning agents.}
    \label{tab:ammo1_results}
    \centering
    \renewcommand{\arraystretch}{1.25}%
    \resizebox{\linewidth}{!}{%
    \begin{tabular}{|l|cccccccccccc|}
    \hline
    \multirow{2}{*}{} & \multicolumn{3}{c|}{ST} & \multicolumn{3}{c|}{SS} & \multicolumn{3}{c|}{SS NB} & \multicolumn{3}{c|}{PPO-18} \\ \cline{2-13} 
     & W & L & T & W & L & T & W & L & T & W & L & T \\ \hline
    Imitation & 0.648 & 0.254 & 0.098 & 0.104 & 0.790 & 0.106 & 0.024 & 0.732 & \textbf{0.244} & 0.066 & 0.834 & \textbf{0.100} \\ \cline{2-13}
    Agent$_0$ & 0.940 & \textbf{0.000} & 0.060 & 0.616 & 0.122 & 0.262 & 0.464 & \textbf{0.046} & 0.490 & 0.682 & \textbf{0.098} & 0.220 \\ \cline{2-13} 
    Agent$_{20}$ & 0.958 & 0.012 & 0.030 & \textbf{0.640} & \textbf{0.118} & 0.242 & \textbf{0.526} & 0.070 & 0.404 & 0.660 & 0.128 & 0.212 \\ \cline{2-13} 
    Agent$_{40}$ & 0.958 & 0.010 & 0.032 & 0.486 & 0.138 & 0.376 & 0.364 & 0.074 & 0.562 & 0.558 & 0.104 & 0.338 \\ \cline{2-13} 
    Agent$_{60}$ & 0.958 & 0.010 & 0.032 & 0.424 & 0.184 & 0.392 & 0.304 & 0.136 & 0.560 & 0.464 & 0.194 & 0.342 \\ \cline{2-13} 
    Agent$_{focus}$ & \textbf{0.982} & \textbf{0.000} & \textbf{0.018} & 0.546 & 0.134 & 0.320 & 0.510 & 0.098 & 0.392 & 0.562 & 0.124 & 0.314 \\ \cline{2-13} 
    Agent$_{incrm}$ & 0.912 & 0.008 & 0.080 & 0.634 & 0.156 & \textbf{0.210} & 0.520 & 0.076 & 0.404 & \textbf{0.700} & 0.120 & 0.180 \\  \hline
    \textit{average} & \textit{0.951} & \textit{0.001} & \textit{0.042} & \textit{0.558} & \textit{0.142} & \textit{0.300} & \textit{0.448} & \textit{0.083} & \textit{0.468} & \textit{0.604} & \textit{0.128} & \textit{0.268}\\
     \hline
    \end{tabular}
}
\end{table}


We conclude this section by noting that Agent$_{0}$, Agent$_{20}$ and Agent$_{incrm}$ appear to be the most consistent across all opponents. Agent$_0$ has the fewest losses, Agent$_{20}$ has the most wins, and Agent$_{incrm}$ has a balance of both. It thus appears that \textit{the most robust policies are obtained by training against a variety of opponents simultaneously} throughout the training process, maximising consistent exposure to all policies. The specific proportions of opponents in each batch have small effects in the results, but the key takeaway is that every training batch should include a mix of multiple opponent types. In the next section, we delve deeper into this result through ablation studies.

\section{Ablation Studies}

We have conducted a series of experiments by modifying the default parameters of the Pommerman Environment to better understand the robustness of the learnt policies, as described below.

\textbf{[A1] Effect of reward reshaping: }The first key question is whether the reward reshaping as per (\ref{eq:reward}) leads to improved results compared to the original reward structure. Figure \ref{fig:reward_shaping} plots the default environment reward during training with the structured (reshaped) reward, and without it (default reward). We can see that the reshaped reward leads to better learning, both against the ST agent as well as after switching to a mix of all opponents.

\textbf{[A2] Variation in effect of powerups: }In the default setting, each agent starts with $1$ ammo (number of simultaneous bomb placements) and blast strength of $2$. We perturb these initial values: ammo (3, 5, 8) and blast strength (5, 8). It is important to note here that these modifications were not applied to opponents. Table~\ref{tab:ammo_blast_agg_results} shows the average win ratio over 2k games with 500 games played against each of ST, SS, SS-NB and PPO-18 agents. Results suggest that Agent$_{focus}$ seems to be performing better when the number of ammo is increased, while Agent$_0$ seems to perform the best when the blast strength has increased. These results further prove that while different skills are learnt by different curricula, the agents trained against the broadest variety of opponents seem to pick up the most robust skills.

\begin{table}[]
\centering
\caption{Win ratio averaged across $2000$ games played, with different initial ammo and blast strength values.}
\label{tab:ammo_blast_agg_results}

\begin{tabular}{|l|ccc|cc|}
\hline
 & \multicolumn{3}{c|}{Ammo} & \multicolumn{2}{c|}{Blast Strength} \\ \cline{2-6} 
 & \multicolumn{1}{c}{3} & \multicolumn{1}{c}{5} & \multicolumn{1}{c|}{8} & \multicolumn{1}{c}{5} & \multicolumn{1}{c|}{8} \\ \hline
Agent$_0$ & 0.581 & 0.433 & 0.083 & \textbf{0.444} & \textbf{0.231} \\ \hline
Agent$_{20}$ & 0.580 & 0.484 & 0.184 & 0.283 & 0.066 \\ \hline
Agent$_{40}$ & 0.508 & 0.402 & 0.078 & 0.330 & 0.101 \\ \hline
Agent$_{60}$ & 0.455 & 0.401 & 0.107 & 0.275 & 0.075 \\ \hline
Agent$_{focus}$ & \textbf{0.598} & \textbf{0.523} & \textbf{0.235} & 0.316 & 0.081 \\ \hline
Agent$_{incrm}$ & 0.584 & 0.488 & 0.192 & 0.437 & 0.207 \\ \hline
\end{tabular}%

\end{table}

\begin{figure}
	\centering
	\includegraphics[width=0.5\textwidth]{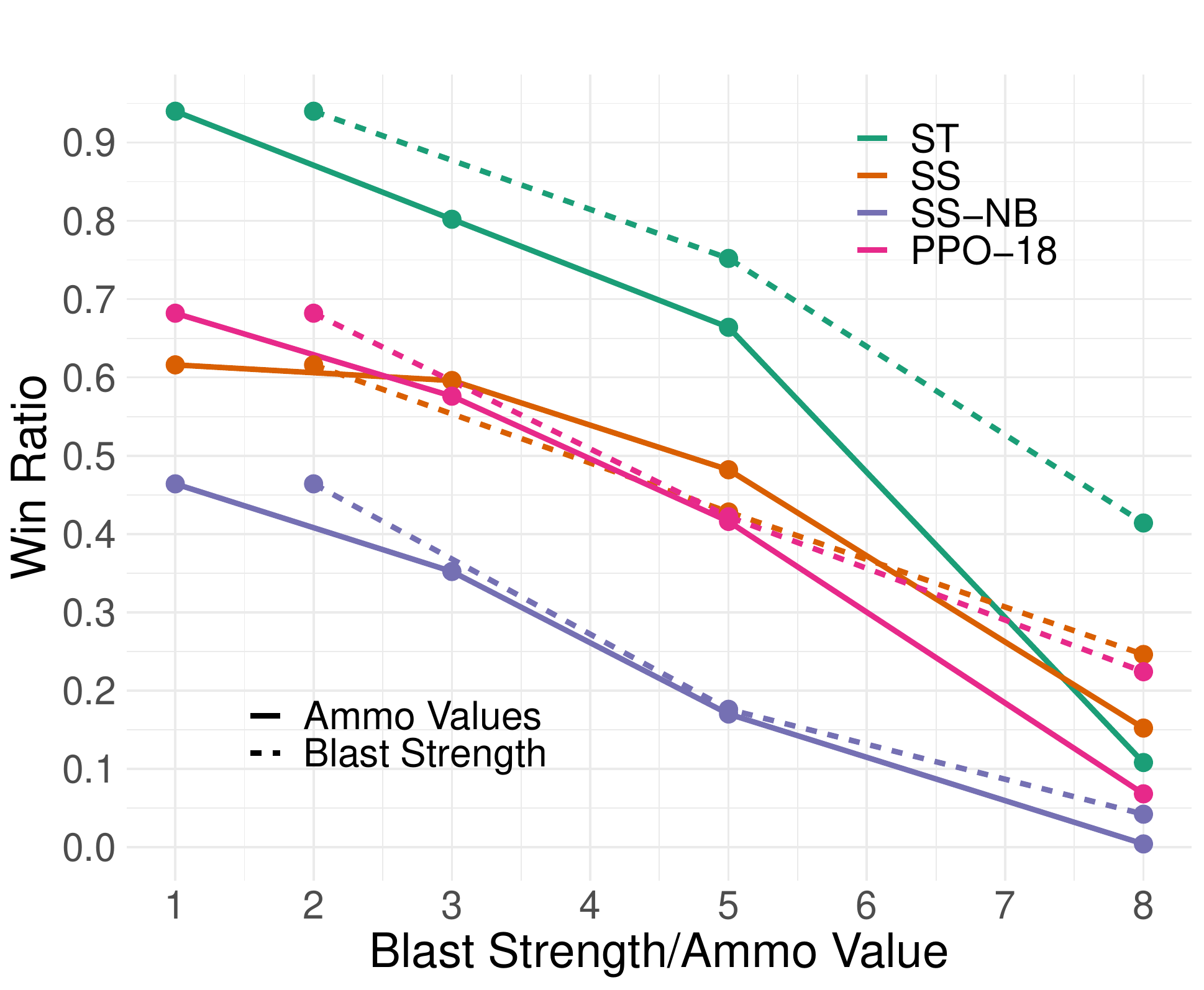}
	\caption{X-axis represents the default values of Ammo / Blast Strength each playing agent gets at the beginning of the game (averaged over total of $2000$ games, $500$ games played against each type of opponent). Y-axis represents winning ratio of Agent$_0$ against different agents which are represented using different colors.}
	\label{fig:ammo_var_win}
\end{figure}

We delve further into this result by doing a fine-grained analysis of Agent$_0$ against all opponents, for different ammo values and blast strengths. show results against different opponents for Agent$_0$. Figure \ref{fig:ammo_var_win} plots the win ratio in all these cases. We conclude that although increasing ammo and blast strength are considered as powerups, including them from the start leads to degradation in performance as there are a lot fewer safe cells for the agent to be in. When the extreme values (like 8) of ammo and blast strength are used, the playing agents are unable to perform adequately in comparison to the normal playing setting. In fact, visual inspections show that when the ammo value is greater than 1 the agent keeps bombs aggressively during the starting stage of the game leading to many suicidal scenarios. 


\begin{figure}[h]
	\begin{subfigure}{.5\textwidth}
    \centering
    \includegraphics[width=0.95\textwidth]{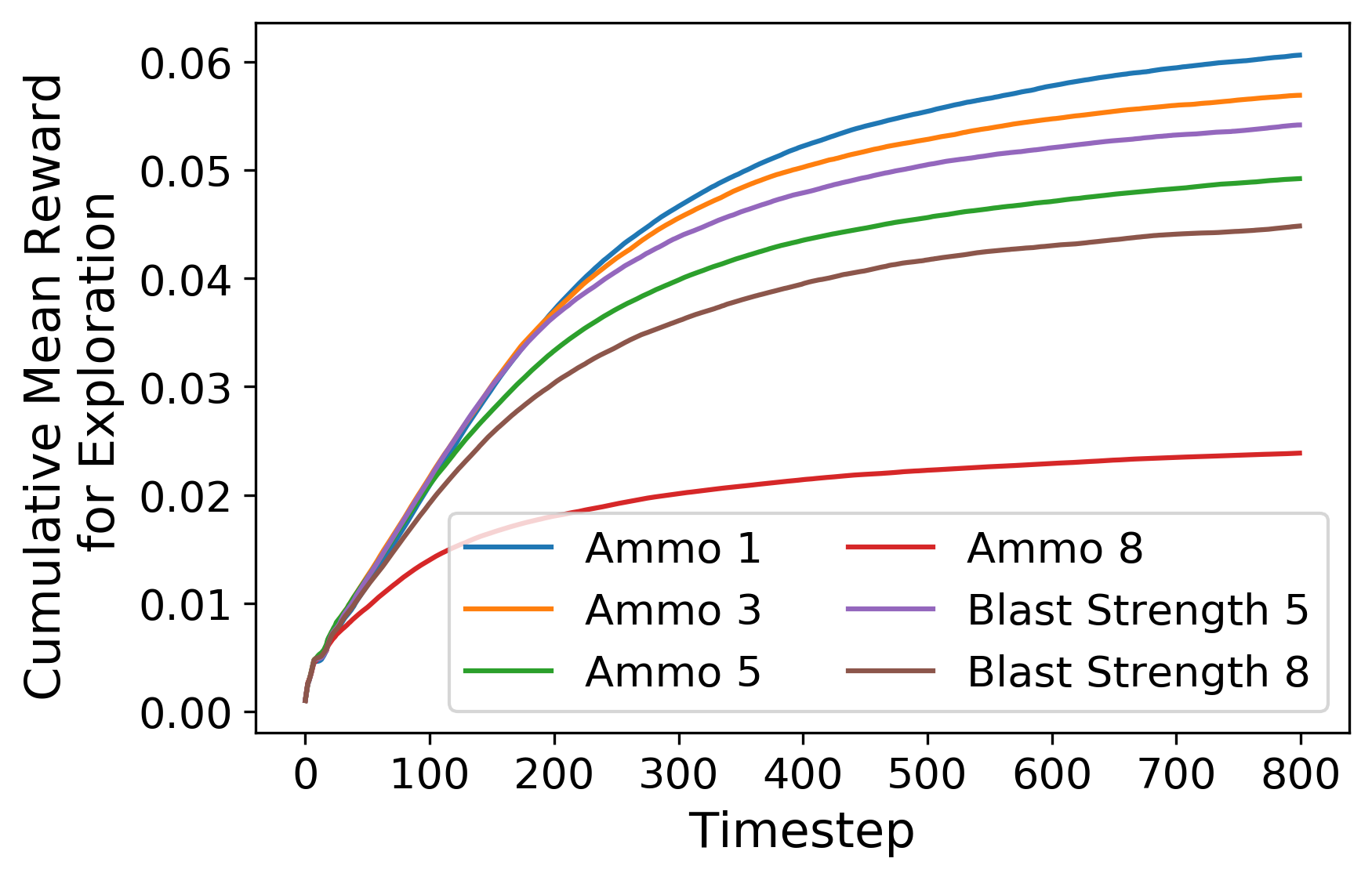}
    \caption{Agent$_0$ exploration reward for different environment settings}
    \label{fig:agent_0K_exploration_rwd}
	\end{subfigure}%
	\begin{subfigure}{.5\textwidth}
		\centering
		\includegraphics[width=0.95\textwidth]{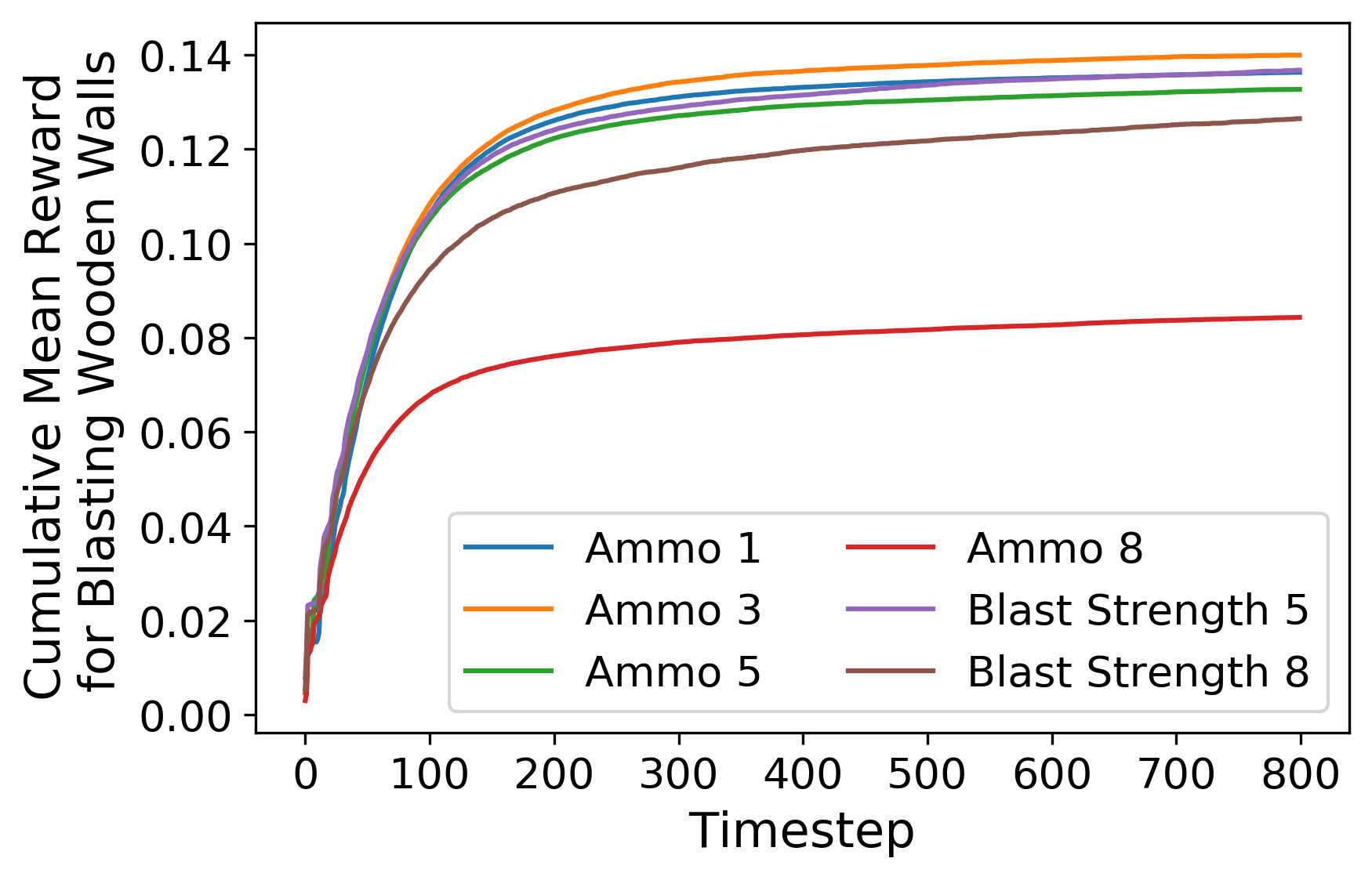}
		\caption{Agent$_0$ Blasting wooden walls reward for different environment settings}
		\label{fig:agent_noch_blasting_walls_rwd}
	\end{subfigure}%
\end{figure}

\begin{figure}
	\begin{subfigure}{.5\textwidth}
    \centering
    \includegraphics[width=0.95\textwidth]{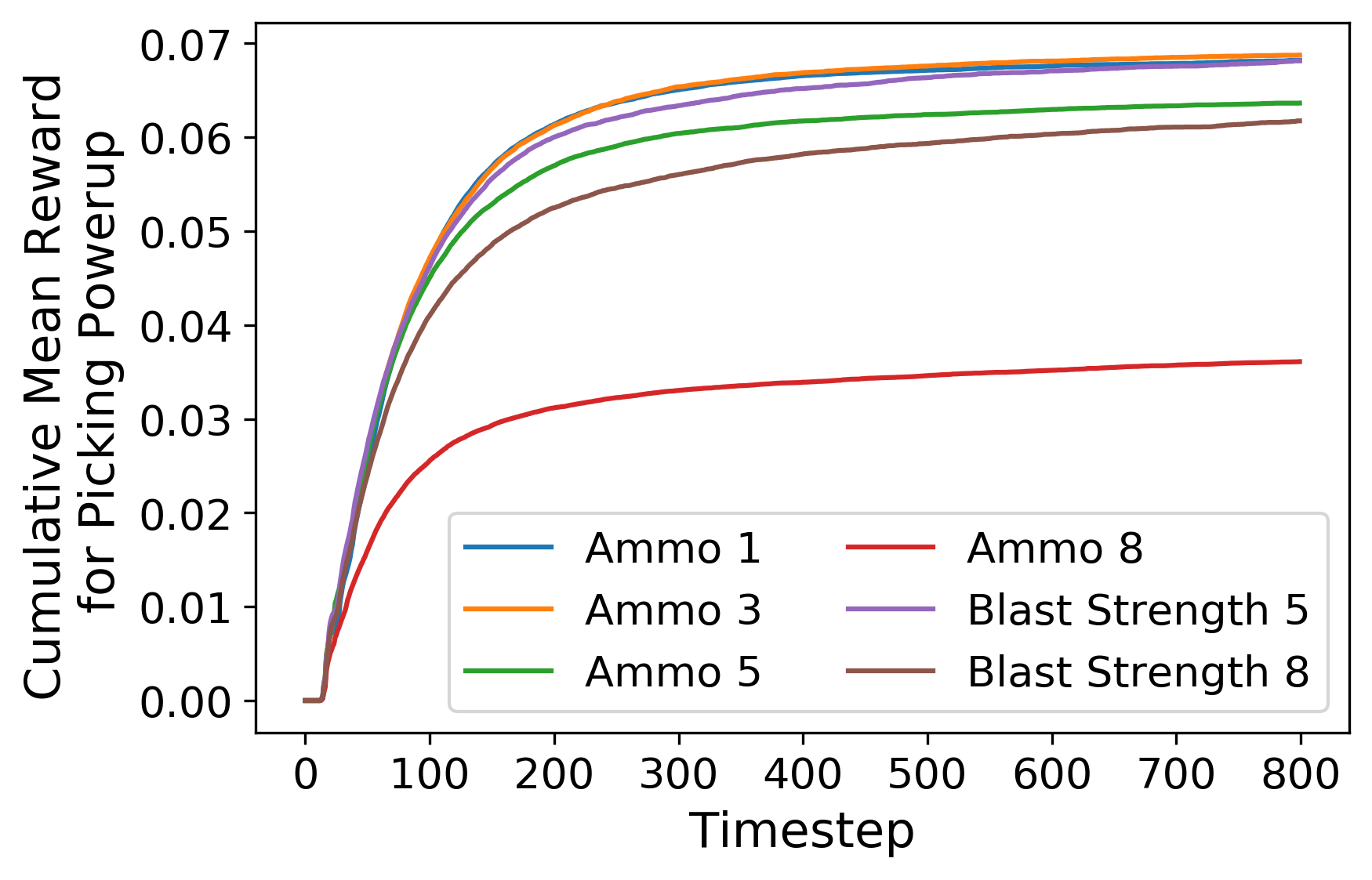}
    \caption{Agent$_0$ picking powerups reward for different environment settings}
    \label{fig:agent_noch_powerup_rwd}
	\end{subfigure}%
	\begin{subfigure}{.5\textwidth}
		\centering
		\includegraphics[width=0.95\textwidth]{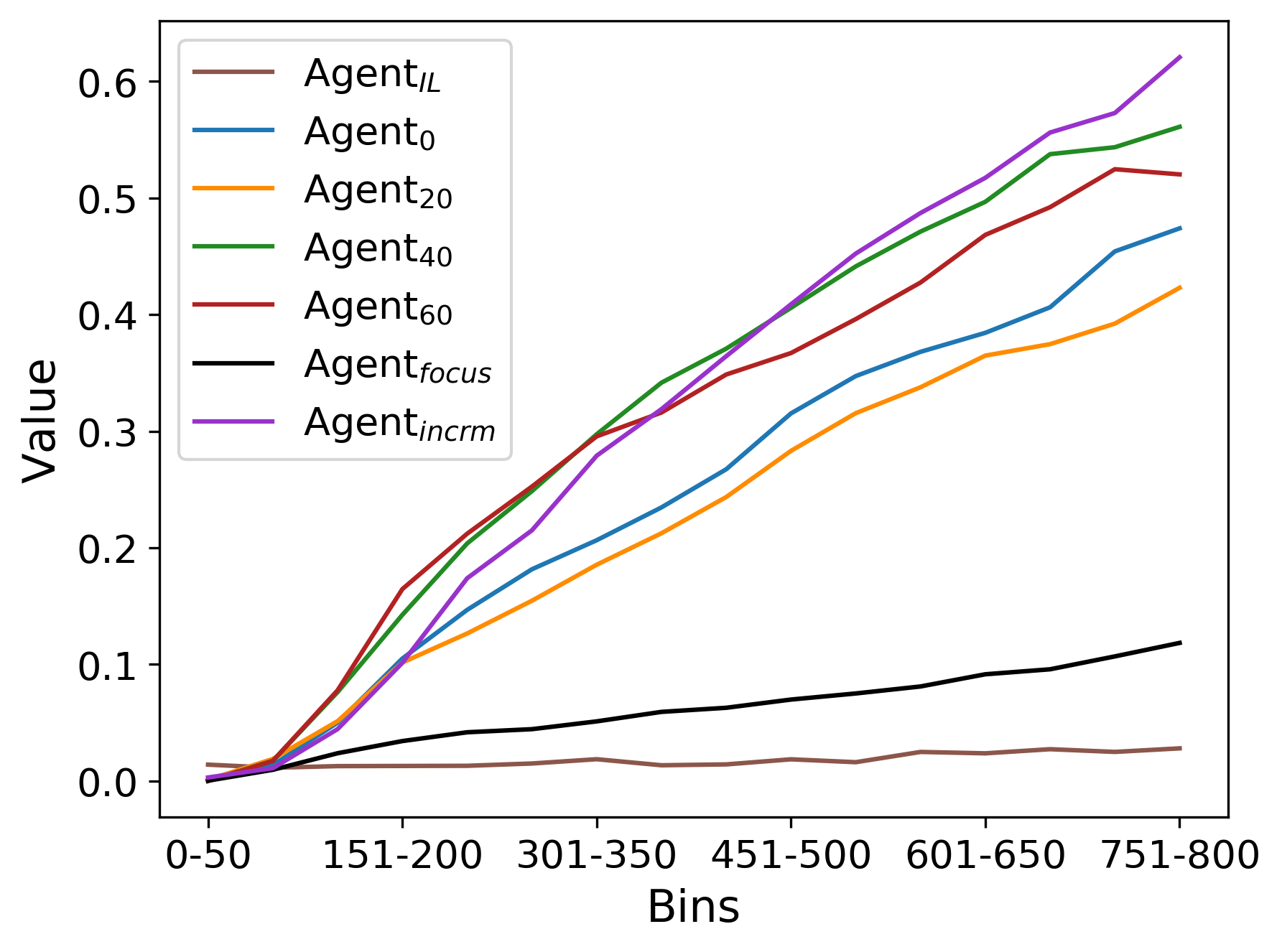}
		\caption{Phenomenon of alternating between adjacent cells. X-axis represents the quantized value of timesteps in with $50$ window, Y-axis is presence/absence averaged across $2000$ games.}
		\label{fig:jitter_detect}
	\end{subfigure}%
\end{figure}

Figures \ref{fig:agent_0K_exploration_rwd}, \ref{fig:agent_noch_blasting_walls_rwd}, and \ref{fig:agent_noch_powerup_rwd} provide insights into how different aspects of intermediate rewards are affected when ammo and blast strength values are changed. Having extreme values of ammo (such as 8) drastically reduces exploration, wooden walls blasting and also powerups. In addition, there is a trend where increasing ammo or blast strength reduces the performance in all the 3 aspects of intermediate rewards. 

Both of these modifications in the default environment settings and the simulated results show that it is difficult for the playing agent to win in such modified scenarios. These modifications can be deemed beneficial by giving the playing agent greater powers compared to the opponents, however it turns out to be the opposite of what is anticipated. In order to play adequately in such settings the agent needs to be trained for these instances from the beginning to play appropriately and be robust to such novel changes. 


\textbf{[A3] Policy jitter: }There is an interesting phenomenon which is observed in Pommerman for learning agents (also reported in~\cite{gao2019skynet, meisheri2019accelerating}), where an agent alternates between two adjacent squares or stays dormant for a longer period of time (typically more than 40 timesteps). Learning agents are particularly susceptible to this behaviour when an enemy agent is near and they both alternate between a fixed pair of cells. Another common trigger for jitter is when there are no immediate goals present in the visible state space (typically later stage of games when all the wooden walls are broken and there is no enemy in sight). Figure~\ref{fig:jitter_detect} shows a graph, where the x-axis is binned values of time steps with $50$ steps per bin, and the y-axis is the fraction of steps when jitter or stationary behaviour is observed in that time window (averaged over 2k games). Agent$_{focus}$ and Agent$_{IL}$ (policy just after imitation learning) seem to be doing much better than other agents. As we enter later stages of the game, there is an increase in jitter across all agents; but the rate of increase is much lower for these two agents. This is a flaw in the policy learnt due to sparse reward and agents often do not break out from such scenarios unless there is a change in their state space leading to the game being tied. The imiation-learnt policy has very little data in the higher values (game lengths are typically very short). Agent$_{focus}$ appears to benefit by playing exclusively against PPO-18 in the final phase of training (this agent does not have a jitter problem).



\section{Conclusion}

In a hybrid cooperative/adversarial multi-agent game such as Pommerman, curriculum learning is a popular way of accelerating training. 
We investigated different curricula for training a robust policy against multiple opponents in a constrained computational budget of 100,000 games, starting from a pre-trained imitation policy. We compared six different sets of curricula varying in the mix and timing of opponent introduction. We found that even though the number of games played against each opponent remains approximately equal during the computational budget, the way opponents are introduced during the training drastically alter the policies. We concluded that having a diverse set of opponents for each gradient step in the policy helps in creating more robust policies. 

We also investigated each policy from different nuances of behaviours, including exploration, policy jitter and perturbation in the environment parameters. We observed that in environments with multiple agents, sparse rewards, and partial observability, focusing only on average rewards could be deceiving. For example, Agent$_{0}$, Agent$_{20}$ and Agent$_{incrm}$ lead to similar expected rewards, but they differ drastically in the exploration of grid, robustness and resilience to ammo and blast variations. In future work, we wish to evaluate the generalisability of these conclusions on other multi-agent environments.

\bibliographystyle{ACM-Reference-Format} 
\bibliography{reference}


\begin{thebibliography}{30}


\ifx \showCODEN    \undefined \def \showCODEN     #1{\unskip}     \fi
\ifx \showDOI      \undefined \def \showDOI       #1{#1}\fi
\ifx \showISBNx    \undefined \def \showISBNx     #1{\unskip}     \fi
\ifx \showISBNxiii \undefined \def \showISBNxiii  #1{\unskip}     \fi
\ifx \showISSN     \undefined \def \showISSN      #1{\unskip}     \fi
\ifx \showLCCN     \undefined \def \showLCCN      #1{\unskip}     \fi
\ifx \shownote     \undefined \def \shownote      #1{#1}          \fi
\ifx \showarticletitle \undefined \def \showarticletitle #1{#1}   \fi
\ifx \showURL      \undefined \def \showURL       {\relax}        \fi
\providecommand\bibfield[2]{#2}
\providecommand\bibinfo[2]{#2}
\providecommand\natexlab[1]{#1}
\providecommand\showeprint[2][]{arXiv:#2}

\bibitem[\protect\citeauthoryear{Aytar, Pfaff, Budden, Paine, Wang, and
  de~Freitas}{Aytar et~al\mbox{.}}{2018}]%
        {aytar2018playing}
\bibfield{author}{\bibinfo{person}{Yusuf Aytar}, \bibinfo{person}{Tobias
  Pfaff}, \bibinfo{person}{David Budden}, \bibinfo{person}{Thomas Paine},
  \bibinfo{person}{Ziyu Wang}, {and} \bibinfo{person}{Nando de Freitas}.}
  \bibinfo{year}{2018}\natexlab{}.
\newblock \showarticletitle{Playing hard exploration games by watching
  youtube}. In \bibinfo{booktitle}{\emph{Advances in Neural Information
  Processing Systems}}. \bibinfo{pages}{2930--2941}.
\newblock


\bibitem[\protect\citeauthoryear{Bain and Sammut}{Bain and Sammut}{1995}]%
        {bain1995framework}
\bibfield{author}{\bibinfo{person}{Michael Bain} {and} \bibinfo{person}{Claude
  Sammut}.} \bibinfo{year}{1995}\natexlab{}.
\newblock \showarticletitle{A Framework for Behavioural Cloning.}. In
  \bibinfo{booktitle}{\emph{Machine Intelligence 15}}.
  \bibinfo{pages}{103--129}.
\newblock


\bibitem[\protect\citeauthoryear{Bernstein, Givan, Immerman, and
  Zilberstein}{Bernstein et~al\mbox{.}}{2002}]%
        {bernstein2002complexity}
\bibfield{author}{\bibinfo{person}{Daniel~S Bernstein}, \bibinfo{person}{Robert
  Givan}, \bibinfo{person}{Neil Immerman}, {and} \bibinfo{person}{Shlomo
  Zilberstein}.} \bibinfo{year}{2002}\natexlab{}.
\newblock \showarticletitle{The complexity of decentralized control of Markov
  decision processes}.
\newblock \bibinfo{journal}{\emph{Mathematics of operations research}}
  \bibinfo{volume}{27}, \bibinfo{number}{4} (\bibinfo{year}{2002}),
  \bibinfo{pages}{819--840}.
\newblock


\bibitem[\protect\citeauthoryear{Clark and Amodei}{Clark and Amodei}{2016}]%
        {FaultyReward}
\bibfield{author}{\bibinfo{person}{Jack Clark} {and} \bibinfo{person}{Dario
  Amodei}.} \bibinfo{year}{2016}\natexlab{}.
\newblock \bibinfo{title}{Faulty Reward Functions in the Wild}.
\newblock
  \bibinfo{howpublished}{\url{https://openai.com/blog/faulty-reward-functions/}}.
\newblock


\bibitem[\protect\citeauthoryear{Daum{\'e}, Langford, and Marcu}{Daum{\'e}
  et~al\mbox{.}}{2009}]%
        {daume2009search}
\bibfield{author}{\bibinfo{person}{Hal Daum{\'e}}, \bibinfo{person}{John
  Langford}, {and} \bibinfo{person}{Daniel Marcu}.}
  \bibinfo{year}{2009}\natexlab{}.
\newblock \showarticletitle{Search-based structured prediction}.
\newblock \bibinfo{journal}{\emph{Machine learning}} \bibinfo{volume}{75},
  \bibinfo{number}{3} (\bibinfo{year}{2009}), \bibinfo{pages}{297--325}.
\newblock


\bibitem[\protect\citeauthoryear{Dorigo and Colombetti}{Dorigo and
  Colombetti}{1994}]%
        {dorigo1994robot}
\bibfield{author}{\bibinfo{person}{Marco Dorigo} {and} \bibinfo{person}{Marco
  Colombetti}.} \bibinfo{year}{1994}\natexlab{}.
\newblock \showarticletitle{Robot shaping: Developing autonomous agents through
  learning}.
\newblock \bibinfo{journal}{\emph{Artificial intelligence}}
  \bibinfo{volume}{71}, \bibinfo{number}{2} (\bibinfo{year}{1994}),
  \bibinfo{pages}{321--370}.
\newblock


\bibitem[\protect\citeauthoryear{Gao, Hernandez-Leal, Kartal, and Taylor}{Gao
  et~al\mbox{.}}{2019}]%
        {gao2019skynet}
\bibfield{author}{\bibinfo{person}{Chao Gao}, \bibinfo{person}{Pablo
  Hernandez-Leal}, \bibinfo{person}{Bilal Kartal}, {and}
  \bibinfo{person}{Matthew~E Taylor}.} \bibinfo{year}{2019}\natexlab{}.
\newblock \showarticletitle{Skynet: A top deep RL agent in the inaugural
  pommerman team competition}.
\newblock \bibinfo{journal}{\emph{arXiv preprint arXiv:1905.01360}}
  (\bibinfo{year}{2019}).
\newblock


\bibitem[\protect\citeauthoryear{Hadfield-Menell}{Hadfield-Menell}{2017}]%
        {reward_bair}
\bibfield{author}{\bibinfo{person}{Dylan Hadfield-Menell}.}
  \bibinfo{year}{2017}\natexlab{}.
\newblock \bibinfo{title}{Cooperatively Learning Human Values}.
\newblock
  \bibinfo{howpublished}{\url{https://bair.berkeley.edu/blog/2017/08/17/cooperatively-learning-human-values/}}.
\newblock


\bibitem[\protect\citeauthoryear{Hester, Vecerik, Pietquin, Lanctot, Schaul,
  Piot, Horgan, Quan, Sendonaris, Dulac-Arnold, et~al\mbox{.}}{Hester
  et~al\mbox{.}}{2017}]%
        {hester2017deep}
\bibfield{author}{\bibinfo{person}{Todd Hester}, \bibinfo{person}{Matej
  Vecerik}, \bibinfo{person}{Olivier Pietquin}, \bibinfo{person}{Marc Lanctot},
  \bibinfo{person}{Tom Schaul}, \bibinfo{person}{Bilal Piot},
  \bibinfo{person}{Dan Horgan}, \bibinfo{person}{John Quan},
  \bibinfo{person}{Andrew Sendonaris}, \bibinfo{person}{Gabriel Dulac-Arnold},
  {et~al\mbox{.}}} \bibinfo{year}{2017}\natexlab{}.
\newblock \showarticletitle{Deep q-learning from demonstrations}.
\newblock \bibinfo{journal}{\emph{arXiv preprint arXiv:1704.03732}}
  (\bibinfo{year}{2017}).
\newblock


\bibitem[\protect\citeauthoryear{Ho and Ermon}{Ho and Ermon}{2016}]%
        {ho2016generative}
\bibfield{author}{\bibinfo{person}{Jonathan Ho} {and} \bibinfo{person}{Stefano
  Ermon}.} \bibinfo{year}{2016}\natexlab{}.
\newblock \showarticletitle{Generative adversarial imitation learning}. In
  \bibinfo{booktitle}{\emph{Advances in neural information processing
  systems}}. \bibinfo{pages}{4565--4573}.
\newblock


\bibitem[\protect\citeauthoryear{Kaelbling, Littman, and Cassandra}{Kaelbling
  et~al\mbox{.}}{1998}]%
        {kaelbling1998planning}
\bibfield{author}{\bibinfo{person}{Leslie~Pack Kaelbling},
  \bibinfo{person}{Michael~L Littman}, {and} \bibinfo{person}{Anthony~R
  Cassandra}.} \bibinfo{year}{1998}\natexlab{}.
\newblock \showarticletitle{Planning and acting in partially observable
  stochastic domains}.
\newblock \bibinfo{journal}{\emph{Artificial intelligence}}
  \bibinfo{volume}{101}, \bibinfo{number}{1-2} (\bibinfo{year}{1998}),
  \bibinfo{pages}{99--134}.
\newblock


\bibitem[\protect\citeauthoryear{Lample and Chaplot}{Lample and
  Chaplot}{2017}]%
        {lample2017playing}
\bibfield{author}{\bibinfo{person}{Guillaume Lample} {and}
  \bibinfo{person}{Devendra~Singh Chaplot}.} \bibinfo{year}{2017}\natexlab{}.
\newblock \showarticletitle{Playing FPS games with deep reinforcement
  learning}. In \bibinfo{booktitle}{\emph{Proceedings of the AAAI Conference on
  Artificial Intelligence}}, Vol.~\bibinfo{volume}{31}.
\newblock


\bibitem[\protect\citeauthoryear{Laskey, Staszak, Hsieh, Mahler, Pokorny,
  Dragan, and Goldberg}{Laskey et~al\mbox{.}}{2016}]%
        {laskey2016shiv}
\bibfield{author}{\bibinfo{person}{Michael Laskey}, \bibinfo{person}{Sam
  Staszak}, \bibinfo{person}{Wesley Yu-Shu Hsieh}, \bibinfo{person}{Jeffrey
  Mahler}, \bibinfo{person}{Florian~T Pokorny}, \bibinfo{person}{Anca~D
  Dragan}, {and} \bibinfo{person}{Ken Goldberg}.}
  \bibinfo{year}{2016}\natexlab{}.
\newblock \showarticletitle{Shiv: Reducing supervisor burden in dagger using
  support vectors for efficient learning from demonstrations in high
  dimensional state spaces}. In \bibinfo{booktitle}{\emph{2016 IEEE
  International Conference on Robotics and Automation (ICRA)}}. IEEE,
  \bibinfo{pages}{462--469}.
\newblock


\bibitem[\protect\citeauthoryear{Lerer and Peysakhovich}{Lerer and
  Peysakhovich}{2018}]%
        {lerer2018learning}
\bibfield{author}{\bibinfo{person}{Adam Lerer} {and} \bibinfo{person}{Alexander
  Peysakhovich}.} \bibinfo{year}{2018}\natexlab{}.
\newblock \showarticletitle{Learning social conventions in markov games}.
\newblock \bibinfo{journal}{\emph{arXiv preprint arXiv:1806.10071}}
  (\bibinfo{year}{2018}).
\newblock


\bibitem[\protect\citeauthoryear{Littman}{Littman}{1994}]%
        {littman1994markov}
\bibfield{author}{\bibinfo{person}{Michael~L Littman}.}
  \bibinfo{year}{1994}\natexlab{}.
\newblock \showarticletitle{Markov games as a framework for multi-agent
  reinforcement learning}.
\newblock In \bibinfo{booktitle}{\emph{Machine learning proceedings 1994}}.
  \bibinfo{publisher}{Elsevier}, \bibinfo{pages}{157--163}.
\newblock


\bibitem[\protect\citeauthoryear{Meisheri and Khadilkar}{Meisheri and
  Khadilkar}{2020}]%
        {meisheri2020sample}
\bibfield{author}{\bibinfo{person}{Hardik Meisheri} {and}
  \bibinfo{person}{Harshad Khadilkar}.} \bibinfo{year}{2020}\natexlab{}.
\newblock \showarticletitle{Sample Efficient Training in Multi-Agent
  Adversarial Games with Limited Teammate Communication}.
\newblock \bibinfo{journal}{\emph{arXiv preprint arXiv:2011.00424}}
  (\bibinfo{year}{2020}).
\newblock


\bibitem[\protect\citeauthoryear{Meisheri, Shelke, Verma, and
  Khadilkar}{Meisheri et~al\mbox{.}}{2019}]%
        {meisheri2019accelerating}
\bibfield{author}{\bibinfo{person}{Hardik Meisheri}, \bibinfo{person}{Omkar
  Shelke}, \bibinfo{person}{Richa Verma}, {and} \bibinfo{person}{Harshad
  Khadilkar}.} \bibinfo{year}{2019}\natexlab{}.
\newblock \showarticletitle{Accelerating Training in Pommerman with Imitation
  and Reinforcement Learning}.
\newblock \bibinfo{journal}{\emph{arXiv preprint arXiv:1911.04947}}
  (\bibinfo{year}{2019}).
\newblock


\bibitem[\protect\citeauthoryear{Nair, McGrew, Andrychowicz, Zaremba, and
  Abbeel}{Nair et~al\mbox{.}}{2018}]%
        {nair2018overcoming}
\bibfield{author}{\bibinfo{person}{Ashvin Nair}, \bibinfo{person}{Bob McGrew},
  \bibinfo{person}{Marcin Andrychowicz}, \bibinfo{person}{Wojciech Zaremba},
  {and} \bibinfo{person}{Pieter Abbeel}.} \bibinfo{year}{2018}\natexlab{}.
\newblock \showarticletitle{Overcoming exploration in reinforcement learning
  with demonstrations}. In \bibinfo{booktitle}{\emph{2018 IEEE International
  Conference on Robotics and Automation (ICRA)}}. IEEE,
  \bibinfo{pages}{6292--6299}.
\newblock


\bibitem[\protect\citeauthoryear{Ng, Harada, and Russell}{Ng
  et~al\mbox{.}}{1999}]%
        {ng1999policy}
\bibfield{author}{\bibinfo{person}{Andrew~Y Ng}, \bibinfo{person}{Daishi
  Harada}, {and} \bibinfo{person}{Stuart Russell}.}
  \bibinfo{year}{1999}\natexlab{}.
\newblock \showarticletitle{Policy invariance under reward transformations:
  Theory and application to reward shaping}. In
  \bibinfo{booktitle}{\emph{ICML}}, Vol.~\bibinfo{volume}{99}.
  \bibinfo{pages}{278--287}.
\newblock


\bibitem[\protect\citeauthoryear{OpenAI}{OpenAI}{2018}]%
        {OpenAI_dota}
\bibfield{author}{\bibinfo{person}{OpenAI}.} \bibinfo{year}{2018}\natexlab{}.
\newblock \bibinfo{title}{OpenAI Five}.
\newblock \bibinfo{howpublished}{\url{https://blog.openai.com/openai-five/}}.
\newblock


\bibitem[\protect\citeauthoryear{Osogami and Takahashi}{Osogami and
  Takahashi}{2019}]%
        {osogami2019real}
\bibfield{author}{\bibinfo{person}{Takayuki Osogami} {and}
  \bibinfo{person}{Toshihiro Takahashi}.} \bibinfo{year}{2019}\natexlab{}.
\newblock \showarticletitle{Real-time tree search with pessimistic scenarios}.
\newblock \bibinfo{journal}{\emph{arXiv preprint arXiv:1902.10870}}
  (\bibinfo{year}{2019}).
\newblock


\bibitem[\protect\citeauthoryear{Panait and Luke}{Panait and Luke}{2005}]%
        {panait2005cooperative}
\bibfield{author}{\bibinfo{person}{Liviu Panait} {and} \bibinfo{person}{Sean
  Luke}.} \bibinfo{year}{2005}\natexlab{}.
\newblock \showarticletitle{Cooperative multi-agent learning: The state of the
  art}.
\newblock \bibinfo{journal}{\emph{Autonomous agents and multi-agent systems}}
  \bibinfo{volume}{11}, \bibinfo{number}{3} (\bibinfo{year}{2005}),
  \bibinfo{pages}{387--434}.
\newblock


\bibitem[\protect\citeauthoryear{Peng, Pang, Yuan, and Gao}{Peng
  et~al\mbox{.}}{2018b}]%
        {peng2018continual}
\bibfield{author}{\bibinfo{person}{Peng Peng}, \bibinfo{person}{Liang Pang},
  \bibinfo{person}{Yufeng Yuan}, {and} \bibinfo{person}{Chao Gao}.}
  \bibinfo{year}{2018}\natexlab{b}.
\newblock \showarticletitle{Continual match based training in Pommerman:
  Technical report}.
\newblock \bibinfo{journal}{\emph{arXiv preprint arXiv:1812.07297}}
  (\bibinfo{year}{2018}).
\newblock


\bibitem[\protect\citeauthoryear{Peng, Abbeel, Levine, and van~de Panne}{Peng
  et~al\mbox{.}}{2018a}]%
        {peng2018deepmimic}
\bibfield{author}{\bibinfo{person}{Xue~Bin Peng}, \bibinfo{person}{Pieter
  Abbeel}, \bibinfo{person}{Sergey Levine}, {and} \bibinfo{person}{Michiel
  van~de Panne}.} \bibinfo{year}{2018}\natexlab{a}.
\newblock \showarticletitle{Deepmimic: Example-guided deep reinforcement
  learning of physics-based character skills}.
\newblock \bibinfo{journal}{\emph{ACM Transactions on Graphics (TOG)}}
  \bibinfo{volume}{37}, \bibinfo{number}{4} (\bibinfo{year}{2018}),
  \bibinfo{pages}{1--14}.
\newblock


\bibitem[\protect\citeauthoryear{Resnick, Gao, M{\'a}rton, Osogami, Pang, and
  Takahashi}{Resnick et~al\mbox{.}}{2020}]%
        {pomm_book_chap}
\bibfield{author}{\bibinfo{person}{Cinjon Resnick}, \bibinfo{person}{Chao Gao},
  \bibinfo{person}{G{\"o}r{\"o}g M{\'a}rton}, \bibinfo{person}{Takayuki
  Osogami}, \bibinfo{person}{Liang Pang}, {and} \bibinfo{person}{Toshihiro
  Takahashi}.} \bibinfo{year}{2020}\natexlab{}.
\newblock \showarticletitle{Pommerman {\&} NeurIPS 2018}. In
  \bibinfo{booktitle}{\emph{The NeurIPS '18 Competition}},
  \bibfield{editor}{\bibinfo{person}{Sergio Escalera} {and}
  \bibinfo{person}{Ralf Herbrich}} (Eds.). \bibinfo{publisher}{Springer
  International Publishing}, \bibinfo{address}{Cham}, \bibinfo{pages}{11--36}.
\newblock
\showISBNx{978-3-030-29135-8}


\bibitem[\protect\citeauthoryear{Ross, Gordon, and Bagnell}{Ross
  et~al\mbox{.}}{2011}]%
        {ross2011reduction}
\bibfield{author}{\bibinfo{person}{St{\'e}phane Ross},
  \bibinfo{person}{Geoffrey Gordon}, {and} \bibinfo{person}{Drew Bagnell}.}
  \bibinfo{year}{2011}\natexlab{}.
\newblock \showarticletitle{A reduction of imitation learning and structured
  prediction to no-regret online learning}. In
  \bibinfo{booktitle}{\emph{Proceedings of the fourteenth international
  conference on artificial intelligence and statistics}}.
  \bibinfo{pages}{627--635}.
\newblock


\bibitem[\protect\citeauthoryear{Schulman, Wolski, Dhariwal, Radford, and
  Klimov}{Schulman et~al\mbox{.}}{2017}]%
        {schulman2017proximal}
\bibfield{author}{\bibinfo{person}{John Schulman}, \bibinfo{person}{Filip
  Wolski}, \bibinfo{person}{Prafulla Dhariwal}, \bibinfo{person}{Alec Radford},
  {and} \bibinfo{person}{Oleg Klimov}.} \bibinfo{year}{2017}\natexlab{}.
\newblock \bibinfo{title}{Proximal Policy Optimization Algorithms}.
\newblock
\newblock
\showeprint[arxiv]{1707.06347}~[cs.LG]


\bibitem[\protect\citeauthoryear{Song, Weng, Su, Yan, Zou, and Zhu}{Song
  et~al\mbox{.}}{2019}]%
        {song2019playing}
\bibfield{author}{\bibinfo{person}{Shihong Song}, \bibinfo{person}{Jiayi Weng},
  \bibinfo{person}{Hang Su}, \bibinfo{person}{Dong Yan},
  \bibinfo{person}{Haosheng Zou}, {and} \bibinfo{person}{Jun Zhu}.}
  \bibinfo{year}{2019}\natexlab{}.
\newblock \showarticletitle{Playing FPS Games With Environment-Aware
  Hierarchical Reinforcement Learning.}. In \bibinfo{booktitle}{\emph{IJCAI}}.
  \bibinfo{pages}{3475--3482}.
\newblock


\bibitem[\protect\citeauthoryear{Zhang and Cho}{Zhang and Cho}{2016}]%
        {zhang2016query}
\bibfield{author}{\bibinfo{person}{Jiakai Zhang} {and}
  \bibinfo{person}{Kyunghyun Cho}.} \bibinfo{year}{2016}\natexlab{}.
\newblock \showarticletitle{Query-efficient imitation learning for end-to-end
  autonomous driving}.
\newblock \bibinfo{journal}{\emph{arXiv preprint arXiv:1605.06450}}
  (\bibinfo{year}{2016}).
\newblock


\bibitem[\protect\citeauthoryear{Zhou, Gong, Mugrai, Khalifa, Nealen, and
  Togelius}{Zhou et~al\mbox{.}}{2018}]%
        {zhou2018hybrid}
\bibfield{author}{\bibinfo{person}{Hongwei Zhou}, \bibinfo{person}{Yichen
  Gong}, \bibinfo{person}{Luvneesh Mugrai}, \bibinfo{person}{Ahmed Khalifa},
  \bibinfo{person}{Andy Nealen}, {and} \bibinfo{person}{Julian Togelius}.}
  \bibinfo{year}{2018}\natexlab{}.
\newblock \showarticletitle{A hybrid search agent in pommerman}. In
  \bibinfo{booktitle}{\emph{Proceedings of the 13th International Conference on
  the Foundations of Digital Games}}. \bibinfo{pages}{1--4}.
\newblock


\end{thebibliography}


\end{document}